\documentclass[10pt,twocolumn,letterpaper]{article}

\usepackage{iccv}
\usepackage{times}
\usepackage{epsfig}
\usepackage{graphicx}
\usepackage{amsmath}
\usepackage{amssymb}
\usepackage{array}
\usepackage{amsfonts}
\usepackage{bm}
\usepackage{algorithm}
\usepackage[noend]{algorithmic}
\usepackage{enumitem}
\usepackage{color}
\usepackage{lipsum}
\usepackage{multirow}
\usepackage{threeparttable}
\usepackage{subcaption}
\usepackage{caption}
\newcommand{\PreserveBackslash}[1]{\let\temp=\\#1\let\\=\temp}
\newcolumntype{C}[1]{>{\PreserveBackslash\centering}p{#1}}
\newcolumntype{R}[1]{>{\PreserveBackslash\raggedleft}p{#1}}
\newcolumntype{L}[1]{>{\PreserveBackslash\raggedright}p{#1}}

\usepackage[breaklinks=true,bookmarks=false]{hyperref}

\iccvfinalcopy % *** Uncomment this line for the final submission

\def\iccvPaperID{3260} % *** Enter the ICCV Paper ID here

\ificcvfinal\fi
\begin{document}

%%%%%%%%% TITLE
\title{WSOD$^2$: Learning Bottom-up and Top-down Objectness Distillation for Weakly-supervised Object Detection}

\author{Zhaoyang Zeng$^{1,2}$\thanks{This work was performed when Zhaoyang Zeng was visiting Microsoft Research as a research intern.}, Bei Liu$^3$, Jianlong Fu$^3$, Hongyang Chao$^{1,2}$, Lei Zhang$^3$\\
{\footnotesize $^1$School of Data and Computer Science, Sun Yat-sen University}\\
{\footnotesize $^2$ The Key Laboratory of Machine Intelligence and Advanced Computing (Sun Yat-sen University), Ministry of Education}\\
{\footnotesize $^3$ Microsoft Research}\\
{\tt\footnotesize zengzhy5@mail2.sysu.edu.cn; \{bei.liu, jianf, leizhang\}@microsoft.com; isschhy@mail.sysu.edu.cn}
}

\maketitle

%\thispagestyle{empty}

%%%%%%%%% ABSTRACT
\begin{abstract}
We study on weakly-supervised object detection (WSOD) which plays a vital role in relieving human involvement from object-level annotations. Predominant works integrate region proposal mechanisms with convolutional neural networks (CNN). Although CNN is proficient in extracting discriminative local features, grand challenges still exist to measure the likelihood of a bounding box containing a complete object (i.e., ``objectness''). In this paper, we propose a novel \textbf{WSOD} framework with \textbf{O}bjectness \textbf{D}istillation (i.e., \textbf{WSOD$^2$}) by designing a tailored training mechanism for weakly-supervised object detection. Multiple regression targets are specifically determined by jointly considering bottom-up (BU) and top-down (TD) objectness from low-level measurement and CNN confidences with an adaptive linear combination. As bounding box regression can facilitate a region proposal learning to approach its regression target with high objectness during training, deep objectness representation learned from bottom-up evidences can be gradually distilled into CNN by optimization. We explore different adaptive training curves for BU/TD objectness, and show that the proposed WSOD$^2$ can achieve state-of-the-art results.

%can boost a relative $11.0\%$ and $9.0\%$ gains for mAP evaluation on PASCAL VOC 2007/2012, respectively, compared with state-of-the-art WSOD models under same setting.
\end{abstract}

%%%%%%%%% BODY TEXT
\vspace{-5mm}
\section{Introduction}

The capability of recognizing and localizing objects in an image reveals a deep understanding of visual information, and has attracted many attentions in recent years. Significant progresses have been achieved with the development of convolutional neural network (CNN) \cite{deng2009imagenet,he2016deep,krizhevsky2012imagenet,szegedy2015going}. However, current state-of-the-art object detectors mostly rely on a large scale of training data which requires manually annotated bounding boxes (e.g., PASCAL VOC 2007/2012 \cite{everingham2010pascal}, MS COCO \cite{lin2014microsoft}, Open Images \cite{OpenImages}). To relieve the heavy labeling effort and reduce cost, weakly-supervised object detection paradigm has been proposed by leveraging only image-level annotations \cite{bilen2016weakly,tang2018weakly,zhang2018zigzag,zhang2018w2f}.

\begin{figure}
  \includegraphics[width=\linewidth]{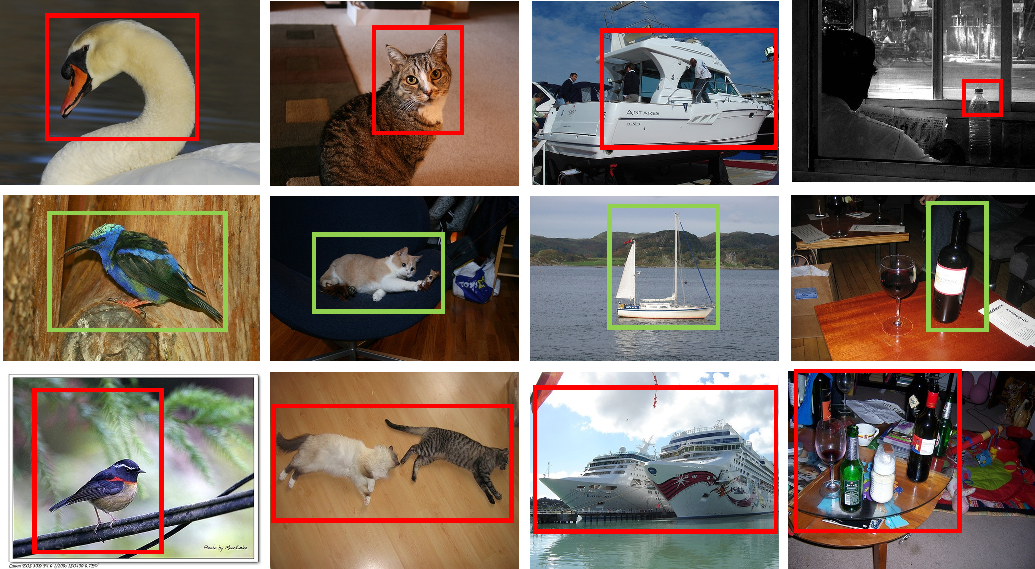}
  \caption{\footnotesize Typical weakly-supervised object detection results produced by OICR \cite{tang2018weakly}. We can observe the partial, correct, and oversized detection results for an object instance in the first, second, and third row, respectively.}
  \label{fig:ob_pt_ct}
  \vspace{-5mm}
\end{figure}

To address weakly-supervised object detection (WSOD) task, most previous works adopt multiple instance learning method to transform WSOD into multi-label classification problems \cite{bilen2016weakly,kantorov2016contextlocnet}. Later on, online instance classifier refinement (OICR) \cite{bai2017multiple} and proposal cluster learning (PCL) \cite{tang2018pcl} are proposed to learn more discriminative instance classifiers by explicitly assigning instance labels. Both OICR and PCL adopt the idea of utilizing the outputs of initial object detector as pseudo ground truths, which has been shown benefits in improving the classification ability of WSOD.
However, a classification model often targets at detecting the existence of objects for a category, while it is not able to predict the location, size and the number of objects in images. This weakness usually results in the detection of partial or oversized bounding boxes, as shown in the first and third rows in Figure~\ref{fig:ob_pt_ct}. %Therefore, there are usually partial or oversized detection results produced by above approaches, as shown in Figure~\ref{fig:ob_pt_ct}. 
The performances of OICR and PCL heavily rely on the accuracy of the initial object detection results, which limit further improvement with large margins. Also, they neglect learning bounding box regression, which plays an important role in the design of modern object detectors \cite{dai2016r,dai2017deformable,he2017mask,lin2017feature,liu2016ssd}. C-WSL integrates bounding box regressors into OICR framework to reduce localization errors, however, it relies on a greedy ground truths selection strategy which requires additional counting annotations \cite{gao2018c}. 

Existing works that rely on the initial weakly-supervised object detection results try to learn the object boundary from feature maps by convolutional neural network (CNN). Although CNN is an expert to learn discriminative local features of an object with image-level labels in a top-down fashion (we call it top-down classifiers in this work), it performs poorly in detecting whether a bounding box contains a complete object without the ground truth for supervision. 

Some low-level feature based object evidences (e.g. color contrast \cite{liu2011learning} and superpixels straddling \cite{alexe2010object}) have been proposed to measure a generic \textit{objectness} that quantifies how likely a bounding box contains an object of any class in a bottom-up way. Inspired by these bottom-up object evidences, in this work, we explore to use their advantage for improving the capability of a CNN model in capturing objectness in images. We propose to integrate these bottom-up evidences that are good at discovering boundary and CNN with powerful representation ability in a single network. 

We propose a \textbf{WSOD} framework with \textbf{O}bjectness \textbf{D}istillation (WSOD$^2$) to leverage bottom-up object evidences and top-down classification output with a novel training mechanism.
%bounding box regression mechanism.
First, given an input image with thousands of region proposals (e.g., generated by Selective Search \cite{uijlings2013selective}), we learn several instance classifiers to predict classification probabilities of each region proposal. Each of these classifiers can help to select multiple high-confident bounding boxes as possible object instances (i.e., pseudo classification and bounding box regression ground truths). Second, we incorporate a bounding box regressor to fine-tune the location and size of each proposal. Third, as each bounding box cannot capture precise object boundaries by CNN features alone, we combine bottom-up object evidences and top-down CNN confidence scores in an adaptive linear combination way to measure the objectness of each candidate bounding box, and assign labels for each region proposal to train the classifiers and regressor.

For some discriminative small bounding boxes that CNN prefers, the bottom-up object evidence (e.g., superpixels straddling) tends to be very low.
WSOD$^2$ can regulate pseudo ground truths to satisfy both higher CNN confidence and low-level object completeness. 
In addition, a bounding box regressor is integrated to reduce the localization error, and augment the effect of bottom-up object evidences during training at the same time.
We design an adaptive training strategy to make the guidance gradually distilled, which enables that a CNN model can be trained strong enough to represent both discriminative local and boundary information of objects when the model converges.

To the best of our knowledge, this work is the first to explore bottom-up object evidences in weakly-supervised object detection task.
The contribution can be summarized as follows:
\begin{enumerate}[nosep]
\item We propose to combine bottom-up object evidences with top-down class confidence scores in weakly-supervised object detection task. 
\item We propose WSOD$^2$ (WSOD with objectness distillation) to distill object boundary knowledge in CNN by a bounding box regressor and an adaptive training mechanism.
\item Our experiments on PASCAL VOC 2007/2012 and MS COCO datasets demonstrate the effectiveness of the proposed WSOD$^2$.
\end{enumerate}

\begin{figure*}[t]

\vspace{-3mm}
\centering
\includegraphics[width=0.95\textwidth]{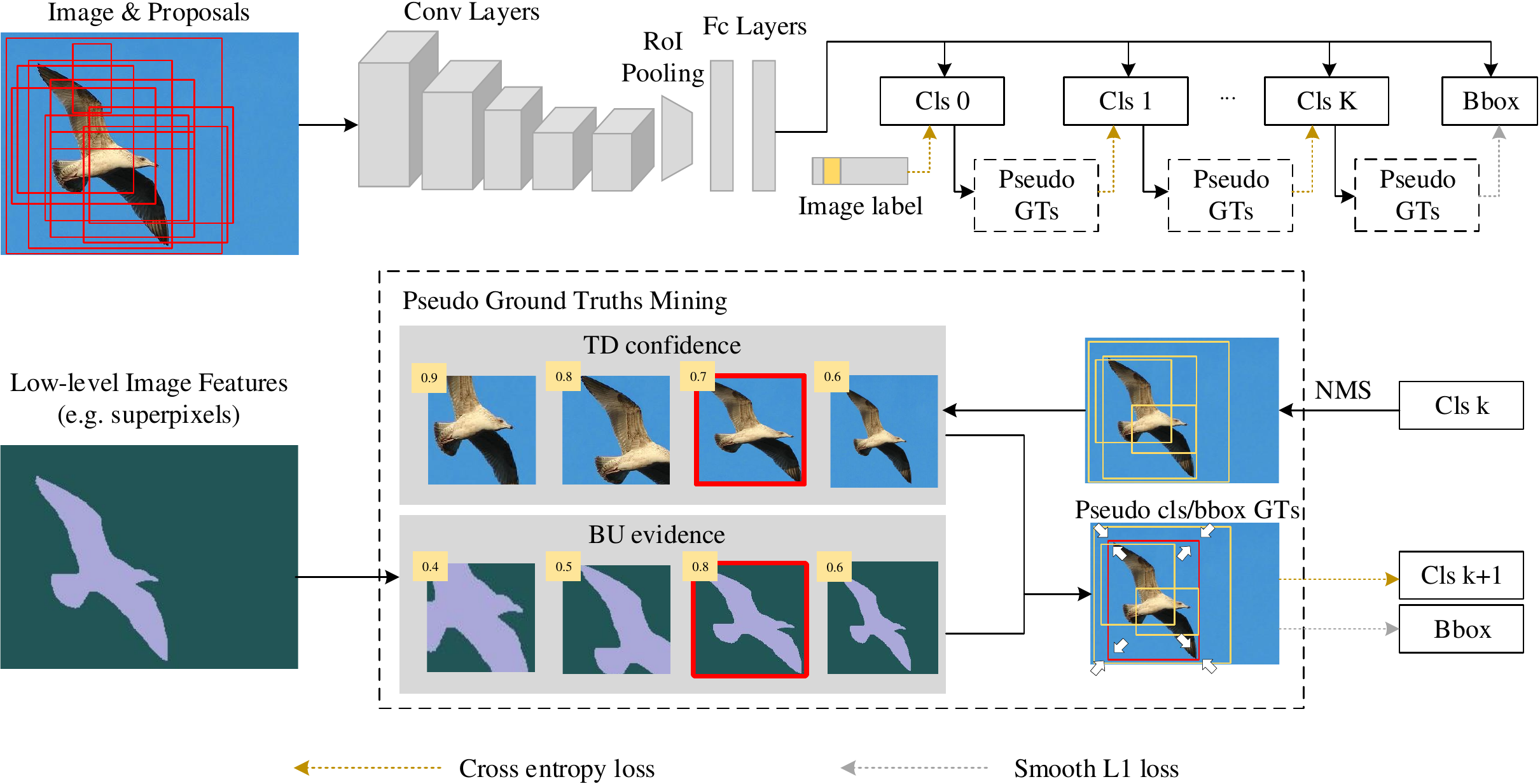}
\caption{\footnotesize The framework of WSOD$^2$. Image with label and pre-computed proposals will be fed into a CNN to obtain region features. The region features will then be passed through several classifiers and a bounding box regressor. Non-maximum suppression (NMS) is applied to mine positive samples from the predictions. Top-down (TD) confidence and bottom-up (BU) evidence are computed by the classification branch and low-level image feature, respectively. They are combined to assign class labels and regression targets for each proposal. ``Cls'' indicates classifier, and ``Bbox'' indicates bounding box regressor. The white arrows indicate the optimization directions for two exemplar region proposals. [Best viewed in color]}
\label{framework}
\vspace{-5mm}
\end{figure*}

%-------------------------------------------------------------------------
\vspace{-3mm}
\section{Related Work}

\subsection{Weakly-supervised Object Detection}
Weakly-supervised object detection has attracted many attentions in recent years. %as it relieves the requirement of expensive and time-consuming human labeling efforts. 
Most existing works adopt the idea of multiple-instance learning \cite{bilen2016weakly,diba2017weakly,jie2017deep,tang2018pcl,bai2017multiple,teh2016attention,wan2018min} to transform weakly-supervised object detection into multi-label classification problems. Bilen \textit{et al.} \cite{bilen2016weakly} proposes WSDDN which performs multiplication on the score of classification and detection branches, so that high-confident positive samples can be selected. Tang \textit{et al.} \cite{tang2018pcl} and Tang \textit{et al.} \cite{bai2017multiple} find that online transforming image-level label into instance-level supervision is an effective way to boost the accuracy, and thus propose to online refine several branches of instance classifiers based on the outputs of previous branches. As class activation map produced by a classifier can roughly localize the object \cite{zhou2016learning, zhu2017soft}, Wei \textit{et al.} \cite{wei2018ts2c} tries to utilize it to generate course detection results, and use them as reference for the later refinement. %Tang \textit{et al.} \cite{bai2017multiple} find that using the outputs of weakly-supervised detectors as pseudo ground truths to train another fully-supervised detector (e.g. Fast/Faster R-CNN \cite{girshick2015fast,ren2015faster}) can always increase the detection performance, and Zhang \textit{et al.}\cite{zhang2018w2f} further designs some selection strategy to select more accurate ones so that the training of second detector can be benefited from them. 
Most previous works rely heavily on pseudo ground truths mining, either online (inside training loop) or offline (after training). Such pseudo ground truths are determined by classification confidence \cite{tang2018pcl,bai2017multiple} or hand-crafted rules \cite{gao2018c,zhang2018w2f}, which are not accurate to measure the objectness of regions. 
% In this paper, we propose to utilize low-level hand-crafted features to extract bottom-up object evidence (e.g., edge density, superpixels straddling), which is capable to evaluate the \textit{objectness} of a region. Such evidence will guide the learning process so that the boundary discovering ability can be inherited to detectors.
\subsection{Bounding Box Regression}
Bounding box regression is proposed in \cite{girshick2014rich}, and is adopted by almost all recent CNN-based fully-supervised object detectors \cite{dai2016r,dai2017deformable,he2017mask,lin2017feature,liu2016ssd} since it can reduce the localization errors of predicted boxes. However, only a few works introduce bounding box into weakly-supervised object detection due to the lack of supervision. Some works consider bounding box regression as a post-processing module. Among which, OICR \cite{bai2017multiple} directly uses the detection results of training set to train Fast R-CNN. W2F \cite{zhang2018w2f} designs some strategies to offline select pseudo ground truth with high precision, based on the output of OICR. Differently, Gao \textit{et al.} \cite{gao2018c} integrate bounding box regressors into OICR inside training loop which leverage addition counting information to help selecting pseudo ground truths.

In this paper, we integrate bounding box regressor into weakly-supervised detector, and assign regression targets by novelly leveraging bottom-up object evidence. 

% Such module not only refines the predicted boxes, but also augments the extracted bottom-up object evidence during training, which can help distillate boundary information to CNN. {\color{red} To the best of our knowledge, we are the first to utilize low-level feature to help bounding boxes discovery in CNN-based weakly supervised object detection.}

\vspace{-3mm}
\section{Approach}

The overview of our proposed \textbf{w}eakly-\textbf{s}upervised \textbf{o}bject \textbf{d}etector with objectness \textbf{d}istillation (\textbf{WSOD$^2$}) is illustrated in Figure~\ref{framework}. We first adopt a based multiple instance detector (i.e. Cls 0) to obtain the initial detected object bounding boxes. Based on the localization of each proposed bounding box, we compute the bottom-up object evidence. Such evidence serves as guidance to transform image-level labels into instance-level supervision. We optimize the whole network in an end-to-end and adaptive fashion. In this section, we will introduce WSOD$^2$ in detail.

\vspace{-2mm}

\subsection{Based Multiple Instance Detector}
\label{TD}
In weakly-supervised object detection, only image-level annotations are available. To better understand semantic information inside an image, we need to go deep into region-level, and analyze the characteristic of each box. We first build a base detector to obtain initial detection result. We follow WSDDN \cite{bilen2016weakly} to adopt the idea of multiple instance learning \cite{tsoumakas2007multi} to optimize the base detector by transforming WSOD into multi-label classification problem. Specifically, given an input image, we first generate region proposals $R$ by Selective Search \cite{uijlings2013selective} and extract region features $\textbf{x}$ by a CNN backbone, an RoI Pooling layer and two fully-connected layers.

Region features $\textbf{x}$ are then fed into two streams by two individual fully-connected layers, and the two produced feature matrices are denoted as $\textbf{x}^c, \textbf{x}^d \in \mathbb{R}^{C\times|R|}$, where $C$ indicates the class number and $|R|$ denotes the proposal number. Two softmax functions are applied on $\textbf{x}^c$ and $\textbf{x}^d$ towards two distinct directions as follows:
\vspace{-2mm}
\begin{equation}
\begin{aligned}
\left[\sigma^c\right]_{ij} = \frac{e^{\left[\textbf{x}^c\right]_{ij}}}{\sum_{k=1}^{C}{e^{\left[\textbf{x}^c\right]_{kj}}}}, 
\left[\sigma^d\right]_{ij} = \frac{e^{\left[\textbf{x}^d\right]_{ij}}}{\sum_{k=1}^{|R|}{e^{\left[\textbf{x}^d\right]_{ik}}}}
\end{aligned},
\vspace{-1mm}
\end{equation}
where $\left[\sigma^c\right]_{ij}$ denotes the prediction of $i^{th}$ class label for $j^{th}$ region proposal, and $\left[\sigma^d\right]_{ij}$ is the weight learned of $j^{th}$ region proposal for $i^{th}$ class. We compute the proposal scores by element-wise product $s = \sigma^c \odot \sigma^d$, and aggregate over the region dimensions to obtain image-level score vector $\phi = [\phi_1, \phi_2,\cdots,\phi_C]$ by $\phi_c = \sum_{r=1}^{|R|}{\left[s\right]_{cr}}$. In such way, we can utilize the image-level class label as supervision and apply binary cross-entropy loss to optimize the base detector. The base loss function is denoted as:
\vspace{-1mm}
\begin{equation}
\begin{aligned}
L_{base} = -\sum_{c=1}^{C}(\hat{\phi_c} log(\phi_c) + (1-\hat{\phi_c})log(1 - \phi_c))
\end{aligned},
\vspace{-1mm}
\end{equation}
where $\hat{\phi_c}=1$ indicates that the input image contains $c^{th}$ class, and $\hat{\phi_c}=0$ otherwise. The prediction score $s$ is considered as initial detection result. However, it is not precise enough and can be further refined as discussed in \cite{bai2017multiple}.

\subsection{Bottom-up and Top-Down Objectness}
\label{BU}
The essence of an object detector is a bounding box ranking function, in which objectness measurement is an important factor. It is common to consider classification confidence as objectness score in recent CNN-based detectors \cite{he2017mask,liu2016ssd,ren2015faster}. However, such strategy has a flaw in weakly-supervised scenario that it is difficult for trained detectors to distinguish complete objects from discriminate object parts or irrelevant background. To relieve this issue, we explore bottom-up object evidences (e.g., superpixels straddling) which play important roles in traditional object detection. 

As stated in \cite{alexe2010object}, objects are standalone things with well-defined boundaries and centers. Thus, we expect a box with a complete object to have a higher objectness score than a partial, oversized or background box. Bottom-up object evidence summarizes the boundary characteristic of common objects, which can help make up for the boundary discovering weakness of CNN.
%guide the CNN learning. 

We propose to integrate bottom-up object evidence to train weakly-supervised object detectors. Specifically, inspired by OICR \cite{bai2017multiple}, we build $K$ instance classifiers on top of $\textbf{x}$, consider the output of $k^{th}$ classifier as the supervision of $(k+1)^{th}$ one, and exploit bottom-up object evidence to guide the network training. Each classifier is implemented by a fully-connected layer and a softmax layer along $C+1$ categories (we consider background as $0^{th}$ class). Formally, for $k^{th}$ classifier, we define the refinement loss function of $k^{th}$ classifier as:
\vspace{-1mm}
\begin{equation}
\label{refinement_loss}
\begin{aligned}
L_{ref}^k = -\frac{1}{|R|}\sum_{r\in R}(w^k_r \cdot CE(p^k_r, \hat{p}^k_r)),
\end{aligned}
\vspace{-1mm}
\end{equation}
where $p^k_r$ denotes the $\{C+1\}$-dim output class probability of proposal $r$, and $\hat{p}^k_r$ indicates its ground truth one-hot label. $CE(p^k_r, \hat{p}^k_r) = -\sum_{c=0}^{C}\hat{p}^k_{rc}log(p^k_{rc})$ is a standard cross entropy function.
Since the real instance-level ground truth labels are unavailable, we use an online strategy to dynamically select pseudo ground truth labels of each proposal in training loop, which will be further explained in Sec~\ref{FD}. We online assign loss weight $w_r^k$ based on the objectness of proposal $r$. Specifically, we first extract bottom-up evidence of $r$ and denote it as $O_{bu}(r)$, then integrate $O_{bu}(r)$ with $O_{td}^k(r)$, which is the class confidence produced by $k^{th}$ classifier. $w_r^k$ is a linear combination of bottom-up evidence and top-down confidence as follows:
\vspace{-1mm}
\begin{equation}
\label{eqn:combination}
\begin{aligned}
w_r^k = \alpha O_{bu}(r) + (1 - \alpha) O_{td}^k(r)
\end{aligned},
\vspace{-1mm}
\end{equation}
where $\alpha$ denotes the impact factor of bottom-up object evidence. Three terms in Eqn~\ref{eqn:combination} are defined as follows:

\noindent \textbf{Bottom-up object evidence $O_{bu}$.} We mainly adopt \textbf{S}uperpixels \textbf{S}traddling(\textbf{SS}) as bottom-up evidence in this work, and we also explore other three evidences: textbf{M}ulti-scale \textbf{S}aliency(\textbf{MS}), \textbf{C}olor \textbf{C}onstrast(\textbf{CC}) and \textbf{E}dge \textbf{D}ensity(\textbf{ED}). Experiment details of these evidences can be found in Sec.~\ref{ablation}.

\noindent \textbf{Top-down Class Confidence $O_{td}$.}
We compute top-down confidence of current branch based on the output of previous branch. Specifically, once we obtain class probability $p^{k-1}_r$ of $(k-1)^{th}$ branch, top-down class confidence of $k^{th}$ branch is computed as:
% \vspace{-2mm}
\begin{equation}
\begin{aligned}
O_{td}^k(r) = \sum_{c=0}^{C}(p^{k-1}_{rc}\cdot\hat{p}^{k}_{rc})
\end{aligned}.
% \vspace{-2mm}
\end{equation}
Since $\hat{p}^{k}$ is a one-hot vector, only one value of $p^{k-1}$ will be picked to computed $O^k_{td}(r)$.

\noindent \textbf{Impact Factor $\alpha$.}
$\alpha$ is the impact factor to balance the effect of bottom-up object evidence and top-down class confidence, which is computed by some weight decay functions. Such design enables boundary knowledge to be distilled into CNN, which will be detailed discussed in Sec.~\ref{FD}.

As bottom-up object evidence and top-down class confidence can measure how likely a box contain a object from the perspective of boundary and semantic information, we consider these two representations as bottom-up and top-down objectness, respectively.

\subsection{Bounding Box Regression}
\label{BBOX}
Bottom-up object evidence is capable to discovery object boundary, so we explore how to make it guide the pre-computed bounding boxes updated during training. An intuitive idea is to integrate bounding box regression to refine the positions and sizes of proposals. 

Bounding box regression is a necessary component in typical fully-supervised object detector, as it is able to reduce localization errors. Although bounding box annotations are unavailable in weakly-supervised object detection, some existing works \cite{gao2018c,tang2018pcl,tang2018weakly,zhang2018w2f} shows that online or offline mining pseudo ground truths and regressing them can boost the performance a lot. Inspired by this idea, we integrate a bounding box regressor on the top of $\textbf{x}$, and make it can be online updated. The bounding box regressor has the same formulation as in Fast R-CNN \cite{girshick2015fast}. For region proposal $r$, the regressor predicts offsets of locations and sizes $t_r=(t^x_r, t^y_r, t^w_r, t^h_r)$, and is further optimized as follows:
% \vspace{-2mm}
\begin{equation}
\begin{aligned}
L_{box} =\frac{1}{|R_{pos}|}\sum_{r=1}^{|R_{pos}|}(w_r^K \cdot smooth_{L1}(t_r, \hat{t}_r))
\end{aligned},
% \vspace{-2mm}
\end{equation}
where $\hat{t}_r$ is computed by the coordinates and sizes difference between $r$ and $\hat{r}$ as described in \cite{girshick2014rich}, where $\hat{r}$ indicates the regression reference. $R_{pos}$ indicates positive (non-background) regions, which will be explained in Sec.~\ref{FD}.  $smooth_{L1}$ function is the same function as defined in \cite{ren2015faster}. $w_r^K$ denotes the regression loss weights computed by the last classification branch.
We compute pseudo regression reference $\hat{r}$ based on the influence of $w_r^K$ which evaluates the objectness of a proposal as we stated in Sec.~\ref{BU}:
\vspace{-1mm}
\begin{equation}
\label{eqn:reg}
\begin{aligned}
\hat{r}=\mathop{\arg\max}_{\{m \in M(K, R)|IoU(m, r)>T_{iou}\}}(w_{m}^K)
\end{aligned},
\vspace{-1mm}
\end{equation}
where $M$ is positive sample mining function which will be explained in Sec~\ref{FD}, and $T_{iou}$ is a specific IoU threshold. Eqn~\ref{eqn:reg} enables each positive region sample to approach a nearby box which has the high objectness. %Such bounding box regression training strategy enables the network to distill the object boundary features. 

We adopt bounding box regression to augment the box prediction during training. We update Eqn~\ref{eqn:combination} as:
\vspace{-1mm}
\begin{equation}
\label{eqn:aug}
\begin{aligned}
w_r^k = \alpha O_{bu}(r') + (1 - \alpha) O_{td}^k(r)
\end{aligned},
\vspace{-1mm}
\end{equation}
where $r'$ is $r$ offset by $t_r$. We keep $O^k_{td}(r)$ unchanged because $O^k_{td}$ contains a RoI feature warping operation, which will be affected by bounding box prediction. In this new formulation, the localization of proposals is online updated. The updated boxes may achieve higher objectness, which means more precise and complete regression targets have higher probability to be selected.

%%%%%%%TO ADD

\subsection{Objectness Distillation}
\label{FD}
% We update Eqn~\ref{refinement_loss} into
% \begin{equation}
% \label{rewrite}
% \begin{aligned}
% L_{ref}^k &= -\frac{1}{|R|}\sum_{r \in R}(\alpha O_{bu}(r')CE(p^k_r, \hat{p}^k_r) \\
% &+ (1-\alpha)O_{td}^k(r)CE(p^k_r, \hat{p}^k_r)))
% \end{aligned}.
% \vspace{-1mm}
% \end{equation}
% Since 
% \begin{equation}
% \label{rewrite}
% \begin{aligned}
% \alpha O_{bu}(r')CE(p^k_r, \hat{p}^k_r) = -\sum_{c=0}^{C}((\alpha O_{bu}(r')\hat{p}^k_rc) log(p^k_rc))
% \end{aligned},
% \vspace{-1mm}
% \end{equation}
Eqn~\ref{refinement_loss} has the similar formulation as knowledge distillation \cite{he2019bag,Hinton2015Distilling}, where the external knowledge comes from bottom-up and top-down objectness. Inside which, $\alpha$ is a weight to balance each knowledge. At the beginning of training, top-down classifiers are not reliable enough, so we expect bottom-up evidences to take the dominant place in the combination (i.e. Eqn~\ref{eqn:combination}). With the guidance of bottom-up evidences, the network will try to regulate the confidence distribution of top-down classifiers to comply with bottom-up evidences. We call this process objectness distillation.
 
As the training proceeds, the reliability of $O_{td}$ increases, and $O_{td}$ inherits the boundary decision ability from $O_{bu}$, while it still keeps the semantic understanding ability because of the classification supervision. Therefore, $\alpha$ can gradually move the attention from bottom-up object evidences to top-down CNN confidences. Specifically, $\alpha$ is computed by some weight decay functions. We survey several weight decay functions including polynomial, cosine and constant functions, and we will compare the effectiveness of different functions in Sec~\ref{ablation}. 

Except $\alpha$, to enable objectness distillation, we also need to determine $\hat{p}^k_r$. We want to leverage bottom-up evidences to enhance boundary representation while keep the semantic recognition ability, thus we utilize output from previous branch of classifier to mine positive proposals. 

Given the output from $(k-1)^{th}$ classifier, we mine pseudo ground truths by following steps:
\begin{enumerate}
\item We apply Non-Maximum Suppression (NMS) on $R$ based on class probability $p^{k-1}_r$ of each proposal $r$ using a pre-defined threshold $T_{nms}$. We denote the kept boxes as $R_{keep}$.
\item For each category $c (c>0)$, if $\hat{\phi}_c=1$, we seek all boxes from $R_{keep}$ whose class confidences on category $c$ are greater than another pre-defined threshold $T_{conf}$, and assign these boxes category label $c$. Specially, if no box is selected, we seek the one with highest score. The set of all seek boxes is denoted as $R_{seek}$.
\item For each seed box in $R_{seek}$, we seek all its neighbor boxes in $R$. Here we consider a box is the neighbor of another box if their Intersection over Union (IoU) is greater than a threshold $T_{iou}$. We denote the set of all neighbor boxes as $R_{neighbor}$. All neighbor boxes will be assigned the same class label as their seed boxes. Other non-seed and non-neighbor boxes will be considered as background. We transform the assigned labels to one-hot vector to obtain all $\hat{p}^k_r$. 
\item Finally, we consider the union set of $R_{seek}$ and $R_{neighbor}$ as the positive proposals: $R_{pos} = R_{seek} \cup R_{neighbor}$.
\end{enumerate}
We group the above operations into function $M(k, R)$ which will return the set of positive proposals, as we mentioned in Sec~\ref{BU} and Sec~\ref{BBOX}. By such way, close positive samples will be assigned same category label, while sample with high objectness will receive high weight. Such information will be distilled into CNN by optimization, thus CNN will gradually increase the ability of discovering object boundary.

\subsection{Training and Inference Details}
\label{PSM}

\textbf{Training.} The overall learning target is formulated as:
\vspace{-1mm}
\begin{equation}
\begin{aligned}
L = L_{base} + \lambda_1\sum_{k=1}^{K}L_{ref}^k + \lambda_2L_{box}
\end{aligned},
\vspace{-1mm}
\end{equation}
where $\lambda_1$ and $\lambda_2$ are hyper-parameters to balance loss weights.. We adopt $\lambda_1=1$ and $\lambda_2=0.3$, and follow \cite{bai2017multiple} to set $K=3$. Since the supervision of all $K$ classifiers comes from previous branches, we set $\alpha=0$ in the first $2,000$ iterations for warm-up. When mining pseudo ground truths, typically we follow \cite{zhang2018w2f} to set $T_{nms}=0.3, T_{conf}=0.7, T_{iou}=0.5$.

\textbf{Inference.} Our model have $K$ refinement classifiers and one bounding box regressor. For each predicted box, we follow \cite{bai2017multiple} to average the outputs from all $K$ classifiers to produce the class confidence, and adjust its position and size using the bounding box regressor. Finally, we apply NMS with threshold $0.3$ to remove redundant detected boxes.

\begin{table*}[t]
\centering
\scriptsize
\begin{tabular}{L{1.6cm}|C{0.3cm}C{0.3cm}C{0.3cm}C{0.3cm}C{0.3cm}C{0.3cm}C{0.3cm}C{0.3cm}C{0.3cm}C{0.3cm}C{0.3cm}C{0.3cm}C{0.3cm}C{0.3cm}C{0.4cm}C{0.3cm}C{0.3cm}C{0.3cm}C{0.3cm}C{0.3cm}|C{0.5cm}}
\hline
\textbf{Evidence} & \textbf{aero} & \textbf{bike} & \textbf{bird} & \textbf{boat} & \textbf{bottle} & \textbf{bus} & \textbf{car} & \textbf{cat} & \textbf{chair} & \textbf{cow} & \textbf{table} & \textbf{dog} & \textbf{horse} & \textbf{mbike} & \textbf{person} & \textbf{plant} & \textbf{sheep} & \textbf{sofa} & \textbf{train} & \textbf{tv} & \textbf{mAP} \\
\hline
\hline
N/A & 58.5 & 63.5 & 46.3 & 25.0 & 18.7 & 66.4 & 63.6 & 55.7 & 26.4 & 45.7 & 42.2 & 43.8 & 48.5 & 63.5 & 15.0 & 24.5 & 44.3 & 49.8 & 62.3 & 54.3 & 45.9 \\
\hline
CC & 62.0 & 64.5 & 44.9 & 24.5 & 19.6 & 70.3 & 62.9 & 52.6 & 20.6 & 54.5 & 44.2 & 49.0 & 55.7 & 64.9 & 15.1 & 22.0 & 49.2 & 56.2 & 52.7 & 58.6 & 47.2 \\
\hline
ED & 52.7 & 60.2 & 44.2 & 32.2 & 20.6 & 65.8 & 60.8 & 67.0 & 21.8 & 57.7 & 38.1 & 51.0 & 57.5 & 66.2 & 15.0 & 25.0 & 52.2 & 54.1 & 61.0 & 37.8 & 47.0 \\
\hline
MS & 62.0 & 66.2 & 41.2 & 25.1 & 19.2 & 68.1 & 61.5 & 60.7 & 12.2 & 52.9 & 47.9 & 61.6 & 58.8 & 65.6 & 18.1 & 17.6 & 47.2 & 59.0 & 54.3 & 51.4 & 47.5 \\
\hline
SS & 61.3 & 63.6 & 44.6 & 26.6 & 21.0 & 65.5 & 61.2 & 49.0 & 25.1 & 52.6 & 44.2 & 58.3 & 64.1 & 65.8 & 16.7 & 21.9 & 49.6 & 53.7 & 59.4 & 57.8 & \textbf{48.1} \\
\hline
CC+ED+MS+SS & 59.5 & 57.6 & 43.1 & 29.7 & 19.7 & 65.4 & 59.7 & 68.1 & 21.5 & 57.6 & 45.7 & 50.5 & 58.4 & 64.0 & 14.6 & 17.2 & 50.4 & 61.2 & 64.9 & 50.0 & 47.9 \\
\hline
\end{tabular}
\caption{\footnotesize Ablation experiments on bottom-up object evidences. We integrate each evidence into WSOD$^2$, and report the mean average precision (mAP) of PASCAL VOC 2007 test split. We also combine all evidences by simply average, the result is listed in the last row.}
\label{ablation_experiment}
\vspace{-3mm}
\end{table*}

\section{Experiments}
\vspace{-1mm}
\subsection{Experimental Setup}
\vspace{-1mm}
\textbf{Datasets and evaluation metrics.} We evaluate our approach on three object detection benchmarks: PASCAL VOC 2007 \& 2012 \cite{everingham2010pascal} and MS COCO \cite{lin2014microsoft}. After removing the bounding box annotations provided by these datasets, we only use images and their label information for training. PASCAL VOC 2007 and 2012 consists of $9,962$ and $22,531$ images of $20$ categories, respectively. For PASCAL VOC, we train on \textit{trainval} split ($5,011$ images for 2007 and $11,540$ for 2012), report mean average precision (mAP) on \textit{test} split, and also adopt correct localization (CorLoc) on \textit{trainval} split to measure the localization accuracy. Both two metrics are performed under the condition of $IoU>0.5$ as a standard setting. MS COCO contains $80$ categories. We train on \textit{train2014} split and evaluate on \textit{val2014} split, which consists of $82,783$ and $40,504$ images, respectively. We report $AP@.50$ and $AP@\left[.50:.05:.95\right]$ on \textit{val2014}.

\newcommand{\tabincell}[2]{\begin{tabular}{@{}#1@{}}#2\end{tabular}} 
\begin{figure}[t]
\begin{minipage}[h]{0.45\linewidth}
\centering
\includegraphics[width=\linewidth]{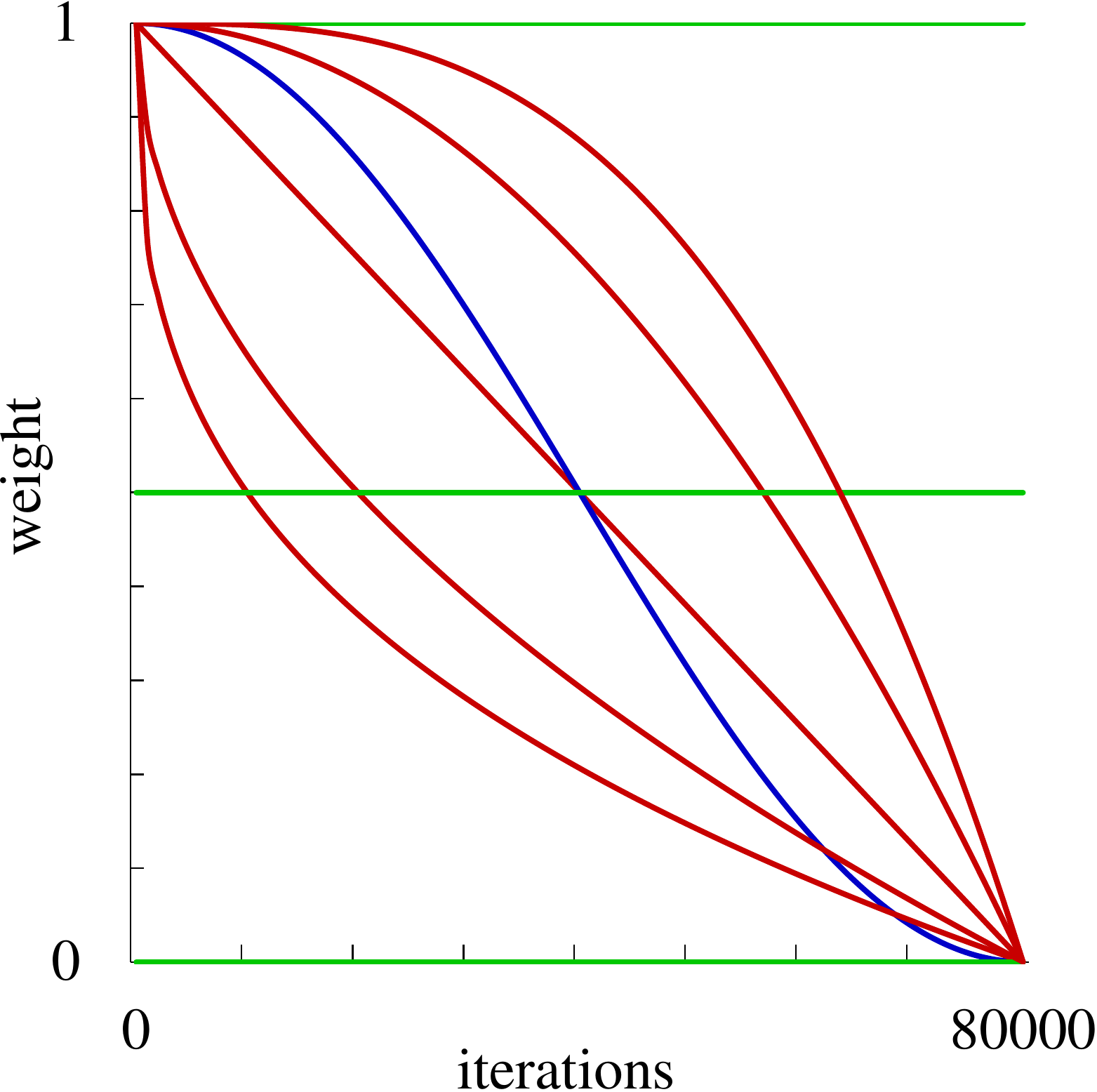}
\subcaption{}
\end{minipage}
\begin{minipage}[h]{0.45\linewidth}
\centering
\scriptsize
\begin{tabular}{c|c|c}
\hline
\textbf{$\alpha$ decay function} & \textbf{$\gamma$} & \textbf{mAP} \\
\hline
\hline
\multirow{3}*{\tabincell{c}{$\alpha=\gamma$ \\  (green curves, from up \\ to down $\gamma=0,\frac{1}{2},1$)}} & 0 & 45.9 \\
\cline{2-3}
~ & $1/2$ & 47.2 \\
\cline{2-3}
~ & 1 & 48.1 \\
\hline
\multirow{5}*{\tabincell{c}{$\alpha=-(\frac{n}{N})^{\gamma} + 1$ \\ (red curves, \\ from up to down \\ $\gamma=3,2,1,\frac{1}{2}, \frac{1}{3}$)}} & 3 & 49.2 \\
\cline{2-3}
~ & 2 & 49.0 \\
\cline{2-3}
~ & 1 & \textbf{50.3} \\
\cline{2-3}
~ & $1/2$ & 46.5 \\
\cline{2-3}
~ & $1/3$ & 46.3 \\
\hline
\multirow{3}*{\tabincell{c}{$\alpha=\frac{(1+cos(\frac{n\pi}{N}))}{2}$ \\ (blue curve)}} & \multirow{3}*{-} & \multirow{3}*{49.7} \\
~ & ~ & ~ \\
~ & ~ & ~ \\
\hline
\end{tabular}
\subcaption{}
\end{minipage}
\vspace{-3mm}
\caption{\footnotesize Ablation study of weight decay functions for $\alpha$. (a) Weight decay curves of different function. (b)mAP of different decay setting on PASCAL VOC 2007 test split. $n$ and $N$ indicate current step and total step number, respectively. [Best viewed in color]}
\label{fig:ablation_alpha}
\vspace{-6mm}
\end{figure}

\textbf{Implementation details.} We adopt VGG16 \cite{simonyan2014very} as the CNN backbone, and use parameters pre-trained on ImageNet \cite{krizhevsky2012imagenet} for initialization. We randomly initialize the weights of all new layers using Gaussian distributions with $0$-mean and standard deviations $0.01$ (except $0.001$ for bounding box regressor), and initialize all new biases to $0$. We follow a widely-used setting \cite{bilen2016weakly,bai2017multiple,tang2018weakly,zhang2018zigzag} to use Selective Search \cite{uijlings2013selective} to generate about $2,000$ proposals for each image. The whole network is end-to-end optimized using SGD with an initial learning rate of $10^{-3}$, weight decay of $0.0005$ and momentum of $0.9$. The overall iteration step number is set to $80,000$ on VOC 2007, and the learning rate will be divided by $10$ at $40,000^{th}$ step. For VOC 2012 we double the iteration step number and learning rate decay step is also doubled to $80,000^{th}$ step. For MS COCO we set iteration step number to $360,000$, and make learning rate decay at $180,000^{th}$ step. We follow \cite{tang2018pcl,bai2017multiple} to adopt multi-scale settings in training. Specifically, the short edge of the input image will be randomly re-scaled to a scale in $\{480, 576, 588, 864, 1280\}$, and we restrict the length of the long edge not greater than $2000$. Besides, horizontal flip of all training images will be also used for training. We report single-scale testing results for ablation study, and report multi-scale testing results when comparing with previous works. All our experiments are implemented based on PyTorch on $4$ NVIDIA P100 GPUs.

\begin{table}[t]
\centering
\scriptsize
\begin{tabular}{c|c|c|c|c}
\hline
\textbf{bbox} & \textbf{NMS} & \textbf{BU} & \textbf{$\alpha$ decay} & \textbf{mAP} \\
\hline
\hline
 & & & & 43.3 \\
\checkmark & & & & 45.1 \\
\checkmark & \checkmark & & & 45.9 \\
\checkmark & \checkmark & \checkmark & & 48.1 \\
\checkmark & \checkmark & \checkmark & \checkmark & 50.3 \\
\hline
\end{tabular}
\vspace{-1mm}
\caption{\footnotesize Ablation study of different components of WSOD$^2$. $\checkmark$ indicates that the component is used. ``NMS'' is unchecked when proposal with highest confidence for each category is used as seed box.
%``bbox'' is checked indicates integrating bounding box regressor. ``bbox aug'' is checked indicates use Eqn~\ref{eqn:aug} instead of Eqn~\ref{eqn:combination}. ``NMS'' is unchecked indicates considering highest-confident proposals as seed boxes rather than applying NMS as in \cite{bai2017multiple}. 
}
\label{tbl:compoment}
\vspace{-4mm}
\end{table}

\begin{table*}[t]
\centering
\scriptsize
\begin{tabular}{L{2.2cm}|C{0.2cm}C{0.2cm}C{0.2cm}C{0.2cm}C{0.3cm}C{0.2cm}C{0.2cm}C{0.2cm}C{0.3cm}C{0.2cm}C{0.3cm}C{0.2cm}C{0.3cm}C{0.4cm}C{0.4cm}C{0.3cm}C{0.3cm}C{0.2cm}C{0.3cm}C{0.3cm}|C{0.7cm}}
\hline
\textbf{method} & \textbf{aero} & \textbf{bike} & \textbf{bird} & \textbf{boat} & \textbf{bottle} & \textbf{bus} & \textbf{car} & \textbf{cat} & \textbf{chair} & \textbf{cow} & \textbf{table} & \textbf{dog} & \textbf{horse} & \textbf{mbike} & \textbf{person} & \textbf{plant} & \textbf{sheep} & \textbf{sofa} & \textbf{train} & \textbf{tv} & \textbf{mAP} \\
\hline
\hline
WSDDN \cite{bilen2016weakly} & 39.4 & 50.1 & 31.5 & 16.3 & 12.6 & 64.5 & 42.8 & 42.6 & 10.1 & 35.7 & 24.9 & 38.2 & 34.4 & 55.6 & 9.4 & 14.7 & 30.2 & 40.7 & 54.7 & 46.9 & 34.8 \\
ContextLocNet \cite{kantorov2016contextlocnet} & 57.1 & 52.0 & 31.5 & 7.6 & 11.5 & 55.0 & 53.1 & 34.1 & 1.7 & 33.1 & 49.2 & 42.0 & 47.3 & 56.6 & 15.3 & 12.8 & 24.8 & 48.9 & 44.4 & 47.8 & 36.3 \\
OICR \cite{bai2017multiple} & 58.0 & 62.4 & 31.1 & 19.4 & 13.0 & 65.1 & 62.2 & 28.4 & 24.8 & 44.7 & 30.6 & 25.3 & 37.8 & 65.5 & 15.7 & 24.1 & 41.7 & 46.9 & 64.3 & 62.6 & 41.2 \\
PCL \cite{tang2018pcl} & 54.4 & 69.0 & 39.3 & 19.2 & 15.7 & 62.9 & 64.4 & 30.0 & 25.1 & 52.5 & 44.4 & 19.6 & 39.3 & 67.7 & 17.8 & 22.9 & 46.6 & 57.5 & 58.6 & 63.0 & 43.5 \\
Tang \textit{et al.} \cite{tang2018weakly} & 57.9 & \textbf{70.5} & 37.8 & 5.7 & 21.0 & 66.1 & \textbf{69.2} & 59.4 & 3.4 & 57.1 & \textbf{57.3} & 35.2 & 64.2 & 68.6 & \textbf{32.8} & \textbf{28.6} & 50.8 & 49.5 & 41.1 & 30.0 & 45.3 \\
C-WSL \cite{gao2018c} & 62.9 & 64.8 & 39.8 & 28.1 & 16.4 & 69.5 & 68.2 & 47.0 & 27.9 & 55.8 & 43.7 & 31.2 & 43.8 & 65.0 & 10.9 & 26.1 & 52.7 & 55.3 & 60.2 & \textbf{66.6} & 46.8 \\
MELM \cite{wan2018min} & 55.6 & 66.9 & 34.2 & 29.1 & 16.4 & 68.8 & 68.1 & 43.0 & 25.0 & \textbf{65.6} & 45.3 & 53.2 & 49.6 & 68.6 & 2.0 & 25.4 & 52.5 & 56.8 & 62.1 & 57.1 & 47.3 \\
ZLDN \cite{zhang2018zigzag} & 55.4 & 68.5 & 50.1 & 16.8 & 20.8 & 62.7 & 66.8 & 56.5 & 2.1 & 57.8 & 47.5 & 40.1 & 69.7 & 68.2 & 21.6 & 27.2 & 53.4 & 56.1 & 52.5 & 58.2 & 47.6 \\
WSCDN \cite{wang2018collaborative} & 61.2 & 66.6 & 48.3 & 26.0 & 15.8 & 66.5 & 65.4 & 53.9 & 24.7 & 61.2 & 46.2 & 53.5 & 48.5 & 66.1 & 12.1 & 22.0 & 49.2 & 53.2 & \textbf{66.2} & 59.4 & 48.3 \\
% W2F\cite{zhang2018w2f} & 63.5 & 70.1 & 50.5 & 31.9 & 14.4 & \textbf{72.0} & 67.8 & \textbf{73.7} & 23.3 & 53.4 & 49.4 & \textbf{65.9} & 57.2 & 67.2 & 27.6 & 23.8 & 51.8 & 58.7 & 64.0 & 62.3 & 52.4 \\ 
\hline
WSOD$^2$ (ours) & \textbf{65.1} & 64.8 & \textbf{57.2} & \textbf{39.2} & \textbf{24.3} & \textbf{69.8} & 66.2 & \textbf{61.0} & \textbf{29.8} & 64.6 & 42.5 & \textbf{60.1} & \textbf{71.2} & \textbf{70.7} & 21.9 & 28.1 & \textbf{58.6} & \textbf{59.7} & 52.2 & 64.8 & \textbf{53.6} \\
WSOD$^2*$ (ours) & 68.2 & 70.7 & 61.5 & 42.3 & 28.0 & 73.4 & 69.3 & 52.3 & 32.7 & 71.9 & 42.8 & 57.9 & 73.8 & 71.4 & 25.5 & 29.2 & 61.6 & 60.9 & 56.5 & 70.7 & 56.0 \\
\hline
\end{tabular}
\vspace{-3mm}
\caption{\footnotesize Mean average precision for different methods on PASCAL VOC 2007 test split. $*$ means training on 07+12 \textit{trainval} splits.}
\label{voc2007test}
\vspace{2mm}

\centering
\scriptsize
\begin{tabular}{L{2.2cm}|C{0.2cm}C{0.2cm}C{0.2cm}C{0.2cm}C{0.3cm}C{0.2cm}C{0.2cm}C{0.2cm}C{0.3cm}C{0.2cm}C{0.3cm}C{0.2cm}C{0.3cm}C{0.3cm}C{0.4cm}C{0.3cm}C{0.3cm}C{0.2cm}C{0.3cm}C{0.3cm}|C{0.7cm}}
\hline
\textbf{method} & \textbf{aero} & \textbf{bike} & \textbf{bird} & \textbf{boat} & \textbf{bottle} & \textbf{bus} & \textbf{car} & \textbf{cat} & \textbf{chair} & \textbf{cow} & \textbf{table} & \textbf{dog} & \textbf{horse} & \textbf{mbike} & \textbf{person} & \textbf{plant} & \textbf{sheep} & \textbf{sofa} & \textbf{train} & \textbf{tv} & \textbf{CorLoc} \\
\hline
\hline
WSDDN \cite{bilen2016weakly} & 65.1 & 58.8 & 58.5 & 33.1 & 39.8 & 68.3 & 60.2 & 59.6 & 34.8 & 64.5 & 30.5 & 43.0 & 56.8 & 82.4 & 25.5 & 41.6 & 61.5 & 55.9 & 65.9 & 63.7 & 53.5 \\
ContextLocNet \cite{kantorov2016contextlocnet} & 83.3 & 68.6 & 54.7 & 23.4 & 18.3 & 73.6 & 74.1 & 54.1 & 8.6 & 65.1 & 47.1 & 59.5 & 67.0 & 83.5 & 35.3 & 39.9 & 67.0 & 49.7 & 63.5 & 65.2 & 55.1 \\
OICR \cite{bai2017multiple} & 81.7 & 80.4 & 48.7 & 49.5 & 32.8 & 81.7 & 85.4 & 40.1 & 40.6 & 79.5 & 35.7 & 33.7 & 60.5 & 88.8 & 21.8 & 57.9 & 76.3 & 59.9 & 75.3 & 81.4 & 60.6 \\
ZLDN \cite{zhang2018zigzag} & 74.0 & 77.8 & 65.2 & 37.0 & \textbf{46.7} & 75.8 & 83.7 & 58.8 & 17.5 & 73.1 & 49.0 & 51.3 & 76.7 & 87.4 & 30.6 & 47.8 & 75.0 & 62.5 & 64.8 & 68.8 & 61.2 \\
PCL \cite{tang2018pcl} & 79.6 & \textbf{85.5} & 62.2 & 47.9 & 37.0 & 83.8 & 83.4 & 43.0 & 38.3 & 80.1 & 50.6 & 30.9 & 57.8 & 90.8 & 27.0 & 58.2 & 75.3 & \textbf{68.5} & \textbf{75.7} & 78.9 & 62.7 \\
C-WSL \cite{gao2018c} & 85.8 & 81.2 & 64.9 & 50.5 & 32.1 & \textbf{84.3} & 85.9 & 54.7 & 43.4 & 80.1 & 42.2 & 42.6 & 60.5 & 90.4 & 13.7 & 57.5 & \textbf{82.5} & 61.8 & 74.1 & \textbf{82.4} & 63.5 \\
Tang \textit{et al.} \cite{tang2018weakly} & 77.5 & 81.2 & 55.3 & 19.7 & 44.3 & 80.2 & \textbf{86.6} & 69.5 & 10.1 & 87.7 & \textbf{68.4} & 52.1 & 84.4 & \textbf{91.6} & \textbf{57.4} & \textbf{63.4} & 77.3 & 58.1 & 57.0 & 53.8 & 63.8 \\
WSCDN \cite{wang2018collaborative} & 85.8 & 80.4 & 73.0 & 42.6 & 36.6 & 79.7 & 82.8 & 66.0 & 34.1 & 78.1 & 36.9 & 68.6 & 72.4 & \textbf{91.6} & 22.2 & 51.3 & 79.4 & 63.7 & 74.5 & 74.6 & 64.7 \\
% W2F \cite{zhang2018w2f} & - & - & - & - & - & - & - & - & - & - & - & - & - & - & - & - & - & - & - & - & \textbf{70.3} \\
\hline
WSOD$^2$ (ours) & \textbf{87.1} & 80.0 & \textbf{74.8} & \textbf{60.1} & 36.6 & 79.2 & 83.8 & \textbf{70.6} & \textbf{43.5} & \textbf{88.4} & 46.0 & \textbf{74.7} & \textbf{87.4} & 90.8 & 44.2 & 52.4 & 81.4 & 61.8 & 67.7 & 79.9 & \textbf{69.5} \\
WSOD$^2*$ (ours) & 89.6 & 82.4 & 79.9 & 63.3 & 40.1 & 82.7 & 85.0 & 62.8 & 45.8 & 89.7 & 52.1 & 70.9 & 88.8 & 91.6 & 37.0 & 56.4 & 85.6 & 64.3 & 74.1 & 85.3 & 71.4 \\
\hline
\end{tabular}
\vspace{-3mm}
\caption{\footnotesize Correct Localization for different methods on PASCAL VOC 2007 trainval split. $*$ means training on 07+12 \textit{trainval} splits.}
\label{voc2007trainval}
\vspace{-3mm}
\end{table*}

\begin{table}[t]
\centering
\scriptsize
\begin{tabular}{c|c|c}
\hline
\textbf{method} & \textbf{mAP} & \textbf{CorLoc} \\
\hline
\hline
OICR \cite{bai2017multiple} & 37.9 & 62.1 \\
PCL \cite{tang2018pcl} & 40.6 & 63.2 \\
Tang \textit{et al.} \cite{tang2018weakly} & 40.8 & 64.9 \\
ZLDN \cite{zhang2018zigzag} & 42.9 & 61.5 \\
WSCDN \cite{wang2018collaborative} & 43.3 & 65.2 \\
% W2F \cite{zhang2018w2f} & \textbf{47.8} & 69.4 \\
\hline
WSOD$^2$ (ours) & \textbf{47.2} \footnotemark[1] & \textbf{71.9} \\
WSOD$^2*$ (ours) & 52.7 \footnotemark[2] & 72.2 \\
\hline
\end{tabular}
\vspace{-3mm}
\caption{\footnotesize Comparisons with different methods on PASCAL VOC 2012 dataset. $*$ means training on 07+12 \textit{trainval} splits.}
\label{voc2012}
\vspace{-5mm}
\end{table}

\subsection{Ablation Study}
\label{ablation}
We conduct ablation studies to demonstrate the effectiveness of WSOD$^2$ on PASCAL VOC 2007.

\textbf{Bottom-up evidences.} 
For bottom-up object evidence, we test the effect of four evidences in both individual and combined ways. The four evidences are list as follows:

1) \textbf{M}ulti-scale \textbf{S}aliency(\textbf{MS}) which summarizes the saliency over several scales; 

2) \textbf{C}olor \textbf{C}onstrast(\textbf{CC}) which computes the color distribution difference with immediate surrounding area; 

3) \textbf{E}dge \textbf{D}ensity(\textbf{ED}) which computes the density of edges in the inner rings;

4) \textbf{S}uperpixels \textbf{S}traddling(\textbf{SS}) which analyzes the straddling of all superpixels.

\noindent Since the value ranges of different evidences are inconsist, we normalize the computed value to $\left[0-1\right]$. For \textbf{CC}, \textbf{ED} and \textbf{MS}, we fix their parameters by setting $\theta^{MS}=0.2, \theta^{CC}=2, \theta^{ED}=2$ empirically due to the lack of supervision. For \textbf{SS}, we follow \cite{felzenszwalb2004efficient} to set $\theta^{SS}_{\sigma}=0.8, \theta^{SS}_k=300$. We refer the readers to \cite{alexe2010object} for more details of these four evidences and the meaning of $\theta^{MS},\theta^{CC},\theta^{ED}, \theta^{SS}$.

To easier analyze the effect of these bottom-up evidences, we simply keep $\alpha=1$ in this ablation experiment for all settings that include these evidences, and $\alpha=0$ for the method that does not involve any bottom-up evidence as a baseline for comparison. We also test the combination of these four evidences by their average. As discussed in \cite{alexe2010object}, linear combination is not a good way to combine them, we conduct this experiment only for evaluating the effectiveness of bottom-up evidences and inspiring future works.

The results are shown in Table~\ref{ablation_experiment}. From the comparison with the baseline, we can find that the performance can increase significantly with the guidance of bottom-up evidences. Table~\ref{ablation_experiment} also includes AP on all categories, from which we find that different evidences may favor different categories. For example, for single evidence, \textbf{ED} favors to ``boat'', while not performs good on ``tv''. Moreover, we can find that this result also agrees with the performance that measures objectness of each evidence as reported in \cite{alexe2010object}, which indicates that these bottom-up evidences are positive correlated to object detection performance. From the result of their combination, we can find it achieves better performance than all single evidences except \textbf{SS}. We believe that linear average is not a correct way to combine these evidence, and better ways can be explored in the future. We adopt \textbf{SS} as bottom-up object evidence in later experiments.

\footnotetext[1]{http://host.robots.ox.ac.uk:8080/anonymous/AVFPZC.html}
\footnotetext[2]{http://host.robots.ox.ac.uk:8080/anonymous/Z4VIWW.html}

\textbf{Impact factor $\alpha$.} We test several weight decay functions, including constant ($\alpha=0, 0.5, 1$), polynomial($\alpha=-(n/N)^{\gamma}+1$, where $\gamma=2,3,1,1/2,1/3$) and cosine ($\alpha=(1 + cos(n\pi /N)/2$) functions where $n$ and $N$ indicate current step and total step number, respectively. The results are shown in Figure~\ref{fig:ablation_alpha}. From the comparison of the first three lines, we find that bottom-up evidences will help the model learn the boundary representation and results in better object detection result. Among different designs, linear decay (i.e., $\alpha=-(n/N)+1$) performs best and the later experiments are conducted based on this setting. We remain exploration of the best parameters for future study.
%There still exist space for hyper-parameters tuning to continue boosting performance, this ablation study only want to show the effectiveness of weight adaptive linear combination.

\textbf{Effect of each component.}
Table~\ref{tbl:compoment} shows the effectiveness of each component. We can find that the bounding box regressor brings at least $2.6$ mAP improvement. Settings that do not use NMS means directly consider the highest-confident box for each category as seed box as OICR \cite{bai2017multiple}. NMS can also improve $0.8$ mAP. Details of bottum-up evidences (BU) and $\alpha$ decay function are discussed above, where both bottom-up evidences and $\alpha$ decay function can bring $2.2$ mAP improvement.

\begin{table}[t]
\centering
\scriptsize
\begin{tabular}{c|c|c}
\hline
\textbf{method} & \textbf{AP@.50} & \textbf{AP@[.50:.05:.95]} \\
\hline
\hline
Ge \textit{et al.} \cite{ge2018multi} & 19.3 & 8.9 \\
PCL \cite{tang2018pcl} & 19.4 & 8.5 \\
PCL + Fast R-CNN \cite{tang2018pcl} & 19.6 & 9.2 \\
\hline
WSOD$^2$ (ours) & \textbf{22.7} & \textbf{10.8} \\
\hline
\end{tabular}
\vspace{-3mm}
\caption{\footnotesize Experiment results of different methods on MS COCO dataset.}
\label{coco}
\vspace{-6mm}
\end{table}

\begin{figure*}[t]
\centering
\includegraphics[width=0.88\linewidth]{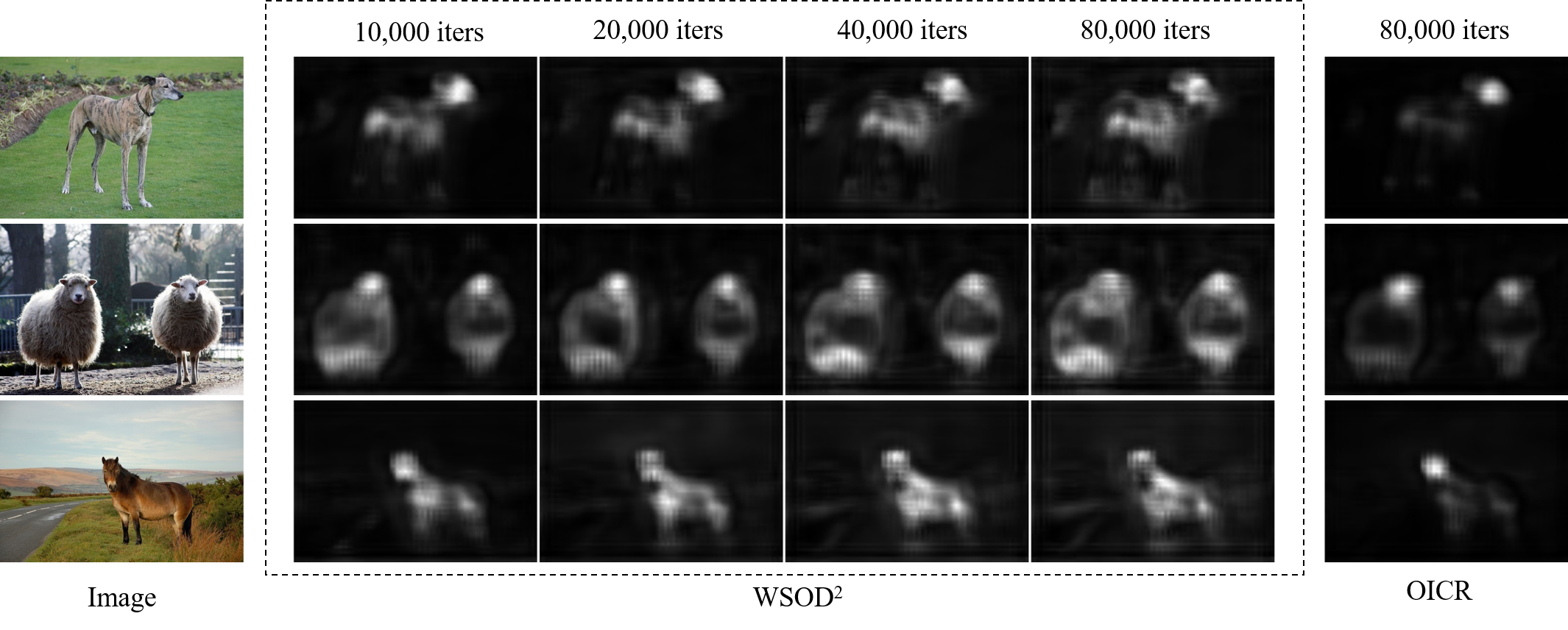}
\vspace{-4mm}
\caption{\footnotesize Visualization of \textit{conv5} feature maps. The response maps are generated by average along all feature map channels, and normalized to $(0, 255)$. The feature maps in middle $4$ columns are extracted by WSOD$^2$ at different iterations. The last column is feature maps we extracted by OICR \cite{bai2017multiple}.}
\label{fig:featuremaps}
\vspace{-4mm}
\end{figure*}

\subsection{Comparisons with State-of-the-Arts}
We evaluate WSOD$^2$ on PASCAL VOC 2007 \& 2012 \cite{everingham2010pascal} and MS COCO datasets \cite{lin2014microsoft}, report the performances and compare with state-of-the-art weakly-supervised detectors. As most of our compared approaches adopt multi-scale testing, we report our multi-scale testing results.

\textbf{AP evaluation on PASCAL VOC.} From Table~\ref{voc2007test} we can find that WSOD$^2$ achieves $53.6$ mAP on PASCAL VOC 2007, which significantly outperforms other end-to-end trainable models \cite{tang2018pcl,bai2017multiple,wang2018collaborative} with at least $5.3$ mAP. WSOD$^2$ is also robust on PASCAL VOC 2012 and achieves $47.2$ mAP, which is shown in Table~\ref{voc2012}.%Compare with approaches that involve several models (e.g. train additional Fast R-CNN) \cite{tang2018pcl,bai2017multiple,tang2018weakly,zhang2018w2f}, WSOD$^2$ still achieve better or comparable performance.

Besides, we follow the common setting in fully-supervised object detection to train WSOD$^2$ on PASCAL VOC 07+12 trainval splits, and denote it as WSOD$^2*$. Such setting achieves a surprising mAP score $56.1$ as shown in the last row of Table\ref{voc2007test}.

\textbf{CorLoc evaluation on PASCAL VOC.} CorLoc evaluates the localization accuracy of detectors on training set. We report results on PASCAL VOC 2007 and 2012 \textit{trainval} split in Table~\ref{voc2007trainval} and Table~\ref{voc2012}, respectively. We can find that WSOD$^2$ significantly surpasses outperforms other end-to-end trainable models \cite{tang2018pcl,bai2017multiple,wang2018collaborative} on both PASCAL VOC 2007 and 2012.

\textbf{AP evaluation on MS COCO.} We report results on MS COCO dataset in Table~\ref{coco}. Since few works report results on MS COCO dataset, we only compare performance with \cite{ge2018multi} and \cite{tang2018pcl}. We can find that WSOD$^2$ outperforms compared works by at least $2$ AP.

\begin{figure}[t]
\centering
\includegraphics[width=\linewidth]{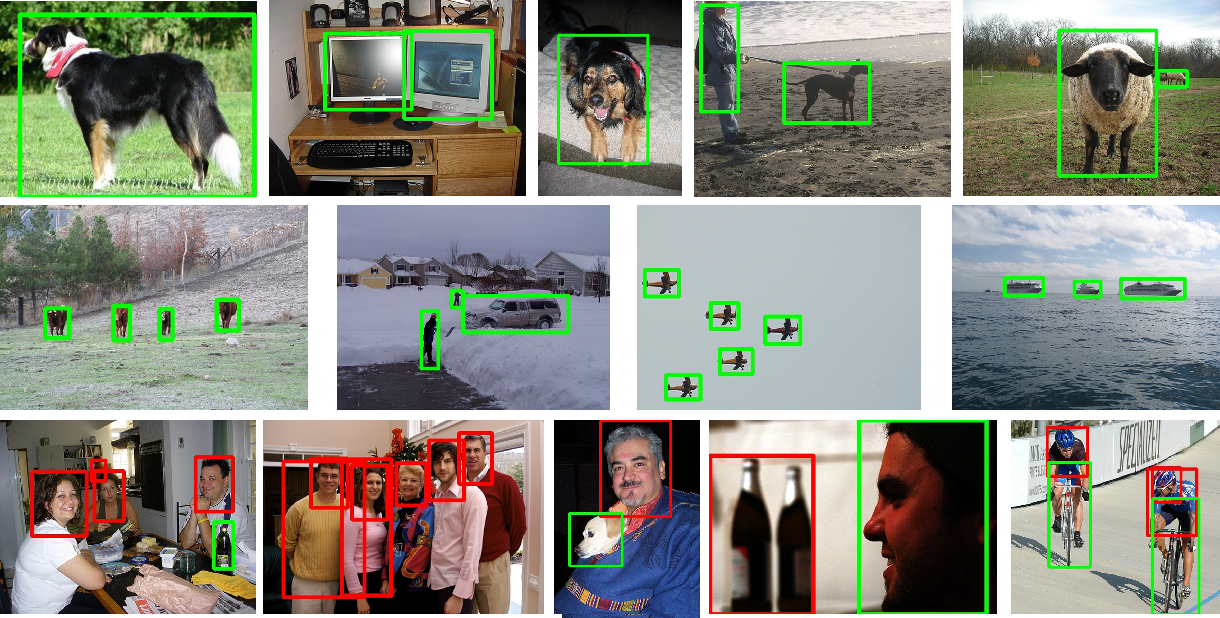}
\vspace{-6mm}
\caption{\footnotesize Example results by WSOD$^2$. Green boxes indicate corrected predictions, and red ones indicate the failure cases. [Best viewed in color]}
\label{fig:experiments}
\vspace{-5mm}
\end{figure}

\subsection{Visualization and Case Study}

We make a qualitative analysis of the effectiveness of WSOD$^2$ compared with OICR. We extract the \textit{conv5} features of trained models, and visualize some cases in Figure~\ref{fig:featuremaps}. The highlighted parts indicate the high response area of the input image in CNN. Compared with OICR, WSOD$^2$ can gradually transfer the response area from discriminate parts to complete objects.

Figure~\ref{fig:experiments} exhibits some successful and failure cases of WSOD$^2$. We obverse that WSOD$^2$ can well handle multiple discrete instances, while there still remains a challenge to solve detection problem in dense scenarios. We also find that for ``person'' class, most weakly-supervised object detectors tend to find human faces. The reason is that in the current datasets, human face is the most common pattern of ``person'', while other parts are often missed in the image. This remains a challenging problem and we can consider leveraging human structure prior in the future.

\vspace{-2mm}
\section{Conclusion}
In this paper, we propose a novel weakly-supervised object detection with bottom-up and top-down objectness distillation (i.e., WSOD$^2$) to improve the deep objectness representation of CNN. Bottom-up object evidence, which could measures the probability of a bounding box including a complete object, is utilized to distill boundary features in CNN in an adaptive training way. We also propose a training strategy that integrates bounding box regression and progressive instance classifier in an end-to-end fashion. We conduct experiments on some standard datasets and settings for WSOD task with our approach. Results demonstrate the effectiveness of our proposed WSOD$^2$ in both quantitative and qualitative way. We also make a thorough analysis on the challenges and possible improvement (e.g., for ``person'' class) of WSOD problem.

\vspace{-2mm}
\section{Acknowledgments}
This work is partially supported by NSF of China under Grant 61672548, U1611461, 61173081, and the Guangzhou Science and Technology Program, China, under Grant 201510010165.

%-------------------------------------------------------------------------
% \subsection{Color}

% Please refer to the author guidelines on the ICCV 2019 web page for a discussion
% of the use of color in your document.

% %------------------------------------------------------------------------
% \section{Final copy}

% You must include your signed IEEE copyright release form when you submit
% your finished paper. We MUST have this form before your paper can be
% published in the proceedings.

{\footnotesize
\bibliographystyle{ieee_fullname}
\bibliography{ms}
}

\end{document}